\documentclass[letterpaper, 10 pt, conference]{ieeeconf}  

\IEEEoverridecommandlockouts                              

\usepackage{epsfig}
\usepackage{graphicx}
\usepackage{amsmath}
\usepackage{amssymb}
\usepackage{booktabs}
\usepackage{times}
\usepackage{mathtools}
\usepackage{float}
\usepackage{xspace}
\usepackage{caption}
\usepackage{subcaption}
\usepackage[inline]{asymptote}
\usepackage{tikz}
\usepackage{makecell}
\usepackage{multirow}
\usepackage{multicol}
\usepackage{booktabs}
\usepackage{pifont}
\usepackage{pgfplots}
\usepackage{nomencl}
\usepackage[english]{babel}
\usepackage[autostyle]{csquotes}
\let\labelindent\relax

\usepackage[inline]{enumitem}
\usepackage{stackengine} 
\usepackage{resizegather}
\usepackage[accsupp]{axessibility}
\usepackage{hyperref}

\usepackage[utf8]{inputenc} 
\usepackage[T1]{fontenc}    
\usepackage{hyperref}       
\usepackage{url}            
\usepackage{booktabs}       
\usepackage{amsfonts}       
\usepackage{nicefrac}       
\usepackage{microtype}      
\usepackage[dvipsnames]{xcolor}         
\usepackage{graphicx}
\usepackage{caption}
\usepackage{amsmath}
\usepackage{amssymb}
\usepackage{mathtools}
\usepackage{wrapfig}
\usepackage{tabularx}
\usepackage{twemojis}
\usepackage{fontawesome5}
\usepackage[utf8]{inputenc} 
\usepackage[T1]{fontenc}    
\usepackage{hyperref}       
\usepackage{url}            
\usepackage{booktabs}       
\usepackage{amsfonts}       
\usepackage{nicefrac}       
\usepackage{microtype}      
\usepackage{multicol}
\usepackage{multirow}
\usepackage{adjustbox}
\usepackage{wrapfig}
\usepackage{changes}
\usepackage{xspace}
\usepackage{float}
\usepackage{cleveref}
\usepackage{colortbl} 
\usepackage{cite}

\newcommand{\eg}[1]{\emph{e.g.,}\xspace}

\newcommand{\filluptopage}[1]{%
  \clearpage
  \loop\ifnum\value{page}<#1\relax
    \null\clearpage
  \repeat
  \loop\ifnum\value{page}=#1\relax
    \null\clearpage
  \repeat
}

\makeatletter
\def\blfootnote{\xdef\@thefnmark{}\@footnotetext}
\makeatother

\newcommand{\shortname}{DreamControl-v2\xspace}

\newcommand{\yes}{\textcolor{green}{\ding{51}}}
\newcommand{\no}{\textcolor{red}{\ding{55}}}

\newcommand{\rom}[1]{\uppercase\expandafter{\romannumeral #1\relax}}

\usepackage{tcolorbox}

\newtcbox{\surface}{on line,
  colback=green!20,      
  colframe=green!50!black, 
  boxrule=1pt,           
  arc=1pt,               
  left=1pt,right=1pt,top=1pt,bottom=1pt  
}

\newtcbox{\reorientation}{on line,
  colback=cyan!20,      
  colframe=cyan!50!black, 
  boxrule=1pt,           
  arc=1pt,               
  left=1pt,right=1pt,top=1pt,bottom=1pt  
}

\newtcbox{\rigid}{on line,
  colback=purple!20,      
  colframe=purple!60!black, 
  boxrule=1pt,           
  arc=1pt,               
  left=1pt,right=1pt,top=1pt,bottom=1pt  
}

\newtcbox{\articulated}{on line,
  colback=yellow!80,      
  colframe=yellow!50!black, 
  boxrule=1pt,           
  arc=1pt,               
  left=1pt,right=1pt,top=1pt,bottom=1pt  
}

\newtcbox{\interactionfree}{on line,
  colback=pink!20,      
  colframe=pink!50!black, 
  boxrule=1pt,           
  arc=1pt,               
  left=1pt,right=1pt,top=1pt,bottom=1pt  
}

\newtcbox{\bimanual}{on line,
  colback=yellow!20,      
  colframe=yellow!50!black, 
  boxrule=1pt,           
  arc=1pt,               
  left=1pt,right=1pt,top=1pt,bottom=1pt  
}

\title{\LARGE \bf
\shortname: Simpler and Scalable Autonomous Humanoid Skills via Trainable Guided Diffusion Priors}

\author{ Sudarshan Harithas$^{1,2}$, Sangkyung Kwak$^1$ , Pushkal Katara$^1$, Srujan Deolasee$^1$, Dvij Kalaria$^{1,3}$, \\ Srinath Sridhar$^2$, Sai Vemprala$^1$, Ashish Kapoor$^1$, Jonathan Chung-Kuan Huang$^1$
\thanks{$^1$General Robotics, USA: Work partially conducted while Sudarshan Harithas and Dvij Kalaria were at General Robotics.}
\thanks{$^2$Brown University, USA.}
\thanks{$^3$University of California, Berkeley, USA.}
}

\begin{document}
\thispagestyle{empty}
\pagestyle{empty}

\makeatletter
\let\@oldmaketitle\@maketitle
\renewcommand{\@maketitle}{%
  \@oldmaketitle
  \centering
  \includegraphics[width=\textwidth]{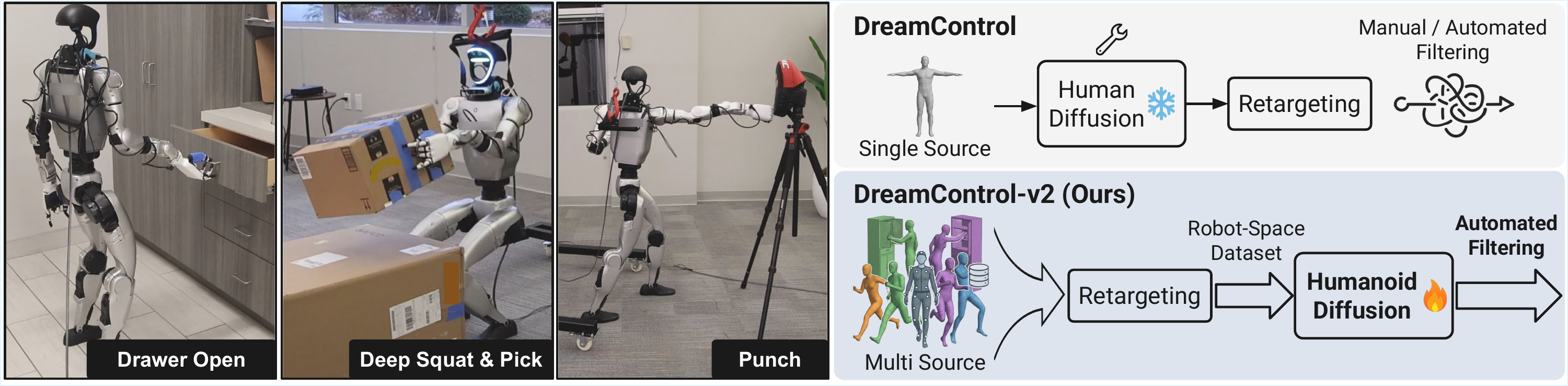}
  \captionof{figure}{%
  \footnotesize{
      \textbf{ DreamControl-v2 enables scalable and autonomous humanoid skill acquisition.} We demonstrate diverse real-world skills (left), including \texttt{Drawer Open}, \texttt{Deep Squat \& Pick }, and \texttt{Punch}. 
      The right panel compares the reference trajectory generation pipelines used before RL: DreamControl relies on a fixed human motion diffusion model, and requires trial-and-error prompting (\faWrench) and manual filtering, whereas \shortname trains a native humanoid diffusion prior from large heterogeneous datasets.
      Retargeting is performed before training, yielding cleaner task-ready trajectories and a pipeline that scales more easily. 
  }}
  \label{teaser}
  \vspace{-4mm}
}
\makeatother

\maketitle
\begin{abstract}
Developing robust autonomous loco-manipulation skills for humanoids remains an open problem in robotics. 
While RL has been applied successfully to legged locomotion, applying it to complex, interaction-rich manipulation tasks is harder given long-horizon planning challenges for manipulation.
A recent approach along these lines is DreamControl, which addresses these issues by leveraging off-the-shelf human motion diffusion models as a generative prior to guide RL policies during training. 
In this paper, we investigate the impact of DreamControl’s motion prior and propose an improved framework that trains a guided diffusion model directly in the humanoid robot’s motion space, aggregating diverse human and robot datasets into a unified embodiment space. 
We demonstrate that our approach captures a wider range of skills due to the larger training data mixture and establishes a more automated pipeline by removing the need for manual filtering interventions. 
Furthermore, we show that scaling the generation of reference trajectories is important for achieving robust downstream RL policies. 
We validate our approach through extensive experiments in simulation and on a real Unitree-G1.  Project website: \url{https://genrobo.github.io/DreamControl-v2/}

\end{abstract}


\section{Introduction}
Recently there have been major advancements in methods for teleoperating humanoid robots~\cite{ze2025twist2,he2024omnih2o,luo2025sonic}, but learning fully autonomous humanoid skills remains a critical challenge, generally requiring a significant amount of data not readily available on the web. 
A promising trend involves utilizing generative priors to produce ``plausible'' robot trajectories (conditioned, e.g., on a text snippet), which are then fed to an Reinforcement Learning (RL) policy to be followed as targets~\cite{tevet2024closd, kalaria2025dreamcontrol}. 
This approach leads to solutions that look human and often transfer more readily to a real robot~\cite{kalaria2025dreamcontrol,tevet2024closd,luo2024omnigrasp}.

A recent contribution along these lines is the DreamControl framework~\cite{kalaria2025dreamcontrol}, which proposed a two stage pipeline for training humanoid policies: \textbf{(1)} first sampling trajectories from a guided diffusion model followed by \textbf{(2)} training an RL policy to track these sampled trajectories (essentially to be teleoperated by these ``dreamed'' trajectories) while simultaneously accomplishing some task of interest autonomously.  
A key benefit of DreamControl is its ability to train policies capable of autonomous environment interactions — leveraging the diffusion model guided by both text and spatio-temporal constraints, allowing for fine-grained control over the timing of contact with various objects in the environment.

Despite some impressive initial results, DreamControl has limited scalability and general utility due to multiple technical issues.
Specifically, DreamControl is built upon a generate-then-retarget pipeline. 
It first uses OmniControl~\cite{xie2024omnicontrol}—a diffusion model trained exclusively on human motion capture (mocap) data—to generate human-centric motion data, which are subsequently retargeted to a specific robot form factor, such as the Unitree-$G1$.
While convenient, this process induces various bottlenecks. 
First, spatial conditioning is enforced in the human body space rather than the robot’s embodiment.
During reference trajectory generation-- for example, given the instruction “pick a box from the ground” and a spatio-temporal constraint $x(t)$ for the hands to approaching and lifting the box, a human motion trajectory is generated through OmniControl~\cite{xie2024omnicontrol} where the human hand follows $x(t)$. 
However, after retargeting this motion to the Unitree G1, the robot’s end-effector no longer follows $x(t)$, but instead arrives at a different position $x_1(t)$. 
The spatial constraint satisfied in human space is therefore not preserved in robot space.
To compensate, practitioners manually adjust the input spatial condition—prompting the human model to reach a different location $x_2(t)$, hoping that after retargeting, the robot will approximate $x(t)$. 
Because the human-motion motion model can generate humans at an arbitrary scale, the parameter $x_2(t)$ must be found through trial-and-error. 
This manual tuning process is labor-intensive and bottlenecks scalability, and is 
especially difficult for long-horizon (tens of seconds), and high-dimensional humanoid loco-manipulation tasks~\cite {kalaria2025dreamcontrol,luo2024omnigrasp}. 
Secondly, models such as OmniControl~\cite{xie2024omnicontrol} can primarily generate AMASS~\cite{AMASS,humanml} style actions, which are in general not relevant for various robotics tasks. 
To address these limitations, DreamControl~\cite{kalaria2025dreamcontrol} proposes several workarounds, including the use of Inverse Kinematics (IK) on tasks not covered within AMASS (such as the \texttt{Drawer Opening} task) and also employs manual filtering of generated trajectories. 
This reliance on task-specific fixes and human intervention inherently restricts the overall scalability of the original DreamControl approach.

To address these limitations, \shortname trains a guided diffusion model directly in the target robot’s action space (e.g., Unitree-$G1$).
Specifically, we construct a robot-space dataset by pre-retargeting a diverse mixture of human motion datasets to the Unitree-$G1$ form factor, and use this dataset to train a flexible robot-motion generation model.
Integrating this model into a DreamControl-inspired workflow streamlines the pipeline and yields two advantages: 
\begin{enumerate*}[label=(\textbf{\arabic*})] 
\item By expanding the training data beyond AMASS~\cite{AMASS}, the model can now capture a diverse range of actions. 
Coupled with the automated filtering process, this approach removes the requirement of task-specific IK and the need for manual trajectory filtering. 
Replacing the out-of-the-box OmniControl model with the \shortname demonstrates scaling and simplicity (Sec.~\ref{exp:task_specific}, Sec.~\ref{sec:perf_analysis}). 
\item  The \shortname diffusion model can now directly generate and accept spatial guidance within the robot's state space, eliminating post-hoc retargeting and the associated trial-and-error tuning (Sec.~\ref{exp:prompt_calib}). 
\end{enumerate*}
Finally, we conduct a detailed analysis of the diffusion model, its effects on downstream RL performance (Sec.~\ref{sec:exp}), and validate our approach with real-world experiments on the Unitree-$G1$.
\vspace{-1mm}

\section{Related Works}
\vspace{-1mm}

\subsection{Motion Synthesis for Robots} 
The success of generative models in domains such as text~\cite{gpt-4,gpt3} and images~\cite{stable-diff} has motivated a range of robotics works that similarly generate actions for manipulation and navigation tasks~\cite{pi0,pi05,kim2024openvla,diffusionpolicy}.
With the availability of large-scale human motion datasets such as AMASS~\cite{AMASS}, motion generation methods have adopted diffusion-based~\cite{mdm,xie2024omnicontrol} and autoregressive~\cite{t2mgpt,motiongpt,harithas2025motionglot} architectures for human motion synthesis.
These approaches generate motion trajectories for a human conditioned on textual descriptions~\cite{mdm,motiongpt,harithas2025motionglot} and spatio-temporal constraints~\cite{xie2024omnicontrol,Pinyoanuntapong2025MaskControl}.
The robustness of these generative modeling techniques has led humanoid robots to adopt similar formulations.
CLOS-D~\cite{tevet2024closd} builds a closely coupled diffusion and RL pipeline to generate physically plausible actions for an animated humanoid.
Similarly, Humanoid-X~\cite{humanoid-x} learns only a text-conditioned model for trajectory generation that is subsequently tracked by a whole-body RL controller. 
In comparison, the \shortname diffusion model can generate trajectories jointly conditioned on both open vocabulary text and spatio-temporal constraints, and performs tasks on a real-robot. 

\subsection{Physics-Based Motion Imitation}

Motion imitation addresses the problem of constructing controllers that enable physics-based tracking of reference motions.
Such motion-tracking formulations provide a general mechanism for acquiring diverse behaviors on humanoid robots.
A large body of work has focused on tracking motions for simulated humanoid characters.
Classical approaches rely on optimization-based controllers~\cite{al2012trajectory} to mimic reference trajectories.
With the advent of learning-based methods, RL has been used to train tracking controllers in simulation~\cite{cheng2024express,ji2024exbody2,he2024learning,he2024omnih2o,twist,ze2025twist2,luo2025sonic}.
More recently, several works have coupled RL with generative priors~\cite{tevet2024closd,luo2024omnigrasp,pan2024synthesizing} for motion tracking and scene interactions within simulation.
However, accurate tracking of a motion reference alone does not constitute autonomous execution.
Methods that require a reference trajectory at inference time are inherently constrained, making them more suitable for teleoperation~\cite{luo2025sonic,ze2025twist2}.
Such reliance on reference trajectories during deployment limits a model’s ability to autonomously adapt to novel environments and task variations.
Inspired by prior work such as OmniGrasp~\cite{luo2024omnigrasp}, both DreamControl and \shortname reformulate motion tracking as a reward signal rather than an input observation.
This design enables policies to execute autonomously at inference time and facilitates direct sim-to-real deployment.

\subsection{Data Sources for Humanoid Loco-Manipulation}
Prior work has explored several data sources for learning loco-manipulation skills on humanoid robots.
One line of research relies on human teleoperation, where skilled operators control the robot to collect high-quality, task-specific demonstrations with real-time feedback~\cite{seo2023deep, he2024omnih2o, twist, ze2025twist2, fu2024humanplus, luo2025sonic}.
While effective for bootstrapping particular tasks, it is inherently expensive and difficult to scale to the breadth of skills required for general-purpose humanoids. 
Another line of work leverages human motion data captured through either Mocap or from RGB videos, primarily for their volume and diversity~\cite{humanoid-x,videomimic,weng2025hdmi,yang2025omniretarget}. 
In contrast to approaches that directly use these motions for physically plausible humanoid control\cite{yang2025omniretarget,weng2025hdmi}, \shortname first trains a generative prior on diverse human datasets, enabling it to synthesize motions spanning a wide range of interactive tasks.
\section{\shortname Methodology}
\renewcommand{\thefigure}{2}
\begin{figure*}[t!]
    \centering
    \includegraphics[
        width=1.\textwidth 
    ]{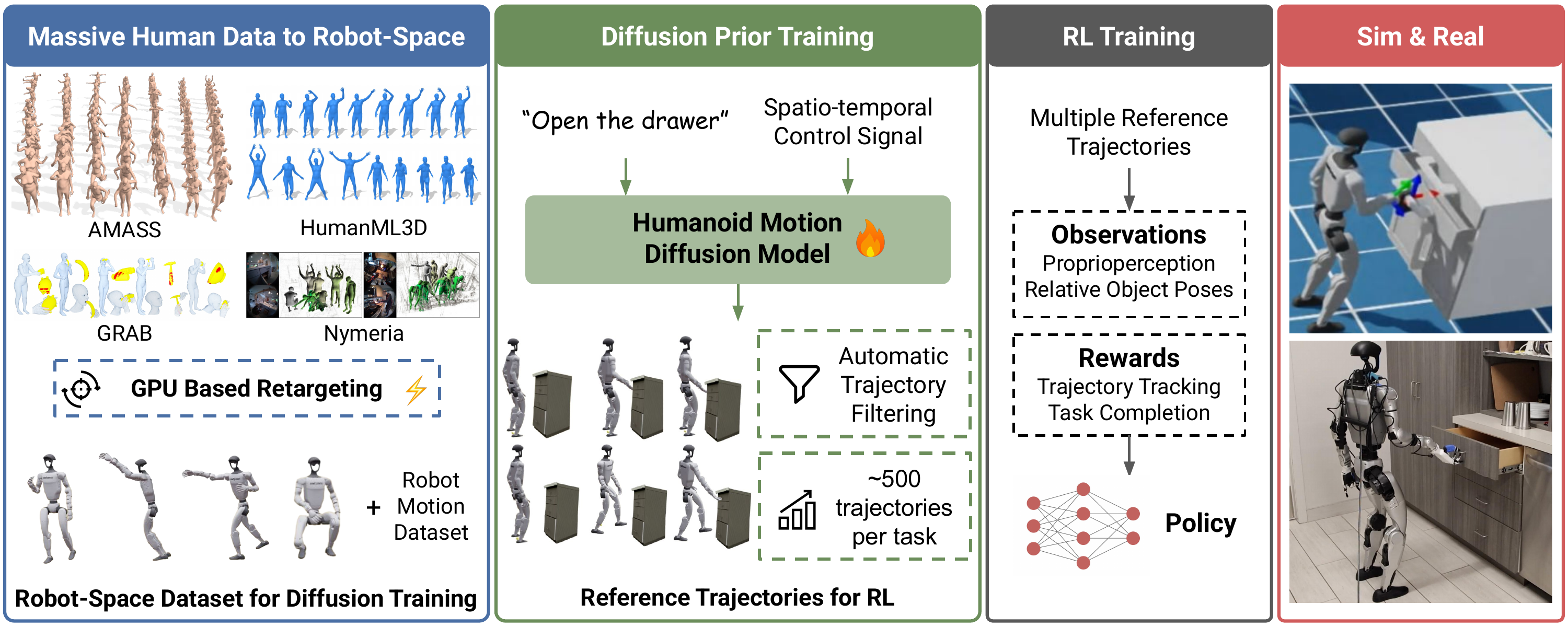}
    \vspace{-0.2in}
    \caption{\footnotesize{\textbf{\shortname Overview.}
    Our four-stage pipeline enables humanoid whole-body manipulation:
    (1) large-scale human motion datasets are retargeted into a unified robot-space;
    (2) a humanoid motion diffusion model is trained to generate reference trajectories conditioned on text and spatio-temporal control signal;
    (3) multiple reference trajectories are used to train a physics-based RL policy;
    (4) The learned policy is deployed in simulation and on a real robot.}}
    \label{fig:method}
    \vspace{-0.2in}
\end{figure*}

We first review the original DreamControl approach, which begins by generating robot trajectories using an off-the-shelf diffusion prior (OmniControl~\cite{xie2024omnicontrol}). 
These trajectories are retargeted to the Unitree-$G1$ embodiment and are constrained to satisfy spatio-temporal requirements, such as making contact with an object at a pre-defined location and time. 
An RL policy is subsequently trained, which receives a reward for both successfully accomplishing the task of interest and for accurately tracking the provided reference trajectory. 

\shortname takes an identical approach to RL and sim-to-real, but differs in the first stage of the pipeline — in this section, we focus on the method for training the diffusion prior to generating humanoid robot trajectories (see Fig.~\ref{fig:method}).


\subsection{Massive Human Data to Robot-Space}\label{sec:robot-space-data}

To train our humanoid motion diffusion model, we construct a large-scale robot-space dataset by retargeting diverse human motion sequences spanning a wide range of activities and interactions into the unified kinematic space of the $G1$ humanoid embodiment. 
This yields a consistent representation suitable for model training.
Conveniently, we can easily train on a union of human mocap datasets and 
newly collected teleoperated data (and thus we have a way to extend the prior to new tasks if they are not already included in the prior).




\subsubsection{Data Sources} 
Our data pool is primarily composed of large-scale human motion datasets AMASS/HumanML3D~\cite{AMASS,humanml}, GRAB~\cite{taheri2020grab}, and Nymeria~\cite{ma2024nymeria} which collectively cover locomotion, everyday activities, and rich human–object interactions.
We also incorporate robot trajectories from OmniRetarget~\cite{yang2025omniretarget}, which originate from human data~\cite{li2023object} but have undergone additional RL and physics-based refinement, resulting in improved bimanual manipulation motions.
These human- and robot-sourced datasets span complementary regions of the motion distribution, enabling our model to learn diverse, whole-body interaction behaviors.


\subsubsection{Human-to-Robot Retargeting}
We retarget all human motion trajectories in our dataset to the $G1$ humanoid. 
The retargeting process begins by selecting a set of keypoints on the human body ($\mathcal{P}_{human}$), specifically 12 joints shared between human and robot, including the hips, knees, ankles, shoulders, elbows, and wrists.
We then solve an optimization problem ($\mathcal{F}_{retarget}$) given in Eq.~\ref{eq:retarget}, that minimizes differences in relative joint angles, relative keypoint positions, foot slip, and a scale factor that compensates for the difference in link lengths between the human body and the $G1$ robot.
The optimization problem results in the joint angles ($q_{robot}$) and global $(\mathtt{SE(3)})$ poses ($T_{robot}$) for the $G1$ at every timestep:

\begin{equation}\label{eq:retarget}
 q_{robot}, T_{robot} =  \mathcal{F}_{retarget}( \mathcal{P}_{human} ).   
\end{equation}
Our retargeting system is built over pyroki~\cite{kim2025pyroki} that uses JAX~\cite{jax2018github} based solvers to perform the optimization. 
We further enhance the efficiency by implementing a single compilation and caching system, along with multi-thread process, which prevents multiple re-compilations overheads.  

\subsubsection{Handling Non-SMPL Parameterizations}
Datasets such as AMASS~\cite{AMASS} and GRAB~\cite{taheri2020grab} represent human motion using the SMPL kinematic model~\cite{SMPL}. 
In contrast, Nymeria~\cite{ma2024nymeria} capture motion with Xsens body suits~\cite{xsens}, producing joint trajectories defined in an alternative kinematic skeleton.
This mismatch, especially in the definition and placement of the pelvis and hip keypoints, introduces inconsistencies between the Xsens and SMPL keypoint sets. 
To resolve these differences, we solve the optimization in Eq.~\ref{eq:xsens_to_smpl} to recover SMPL parameters: the joint angles \(\theta\) and global pose \(T_{\text{pose}} \in \mathrm{SE}(3)\) that best explain the Xsens keypoint observations. 
Here, \(\mathcal{F}_{\text{smpl}}\) denotes the SMPL \emph{differentiable forward kinematics} mapping \((\theta, T_{\text{pose}})\) to 3D SMPL keypoints. 
We employ a confidence-weight vector \(\mathbf{w}\) that assigns higher weights to reliable joints (e.g., wrists, knees, ankles) and lower weights elsewhere. 
The resulting SMPL parameters produce keypoints that are consistent with AMASS and GRAB.
\vspace{-2mm}

\begin{equation}
\label{eq:xsens_to_smpl}
\mathcal{P}_{\text{human}}
= \arg\min_{\theta,\, T_{\text{pose}}}
\left\|
  \mathbf{w} \odot
  \bigl(\mathbf{x}_{\text{xsens}} - \mathcal{F}_{\text{smpl}}(\theta, T_{\text{pose}})\bigr)
\right\|_2^2.
\end{equation}

\subsubsection{Filtering and Pre-processing} \label{sec:nymeria_process}
Not all raw trajectories are suitable for training,   so we apply the feasibility filters from PHC~\cite{Luo2023PerpetualHC,uhc}  to remove motions which are infeasible without environment context (e.g., ``treadmill running'' is not possible to transfer to humanoid without knowing the speed of the belt.).
Nymeria additionally contains text annotations describing atomic actions;   we use a keyword-based retrieval to extract task-specific segments e.g.,   \texttt{``drawer''}, \texttt{``open''}, \texttt{``cabinet''} for articulated object manipulation.
The keyword-based search returns timestamped trajectories that are associated with a particular keyword. 
We segment the trajectory using the timestamps and include it in the dataset.
Lastly, we do not apply any filtering over the GRAB~\cite{taheri2020grab} and OmniRetarget~\cite{yang2025omniretarget} sequences.

\subsection{Diffusion Prior Training}\label{sec:diff-prior-train}
Given the Robot-Space dataset containing $G1$ motion trajectories (from Sec.~\ref{sec:robot-space-data}), we train a diffusion model generating reference trajectories ($\mathcal{T} \in \mathbb{R}^{N \times D}$) conditioned on open vocabulary text instructions $\mathbf{p}$ and spatio-temporal control signals $\mathbf{c} \in \mathbb{R}^{N \times J \times 3}$ over any joint at any time, where $N$ is the length of the trajectory, $J$ is the number of joints, and $D$ is the dimension of the $G1$ pose representation.
The control signal $\mathbf{c}$ provides 3D-control positions per-joints across timestamps (or for a select number of key frames), while for all other joints that are left unconstrained is set to zero.
For example, in the \texttt{Drawer Open} task, the text and spatial controls jointly specify: ``open the drawer'' while constraining the right wrist to reach the drawer handle at position $\mathit{xyz}$ at time $t_j$, the intended interaction moment.
Depending on the task, the spatio-temporal controls may be sparse and provided for only one or two joints, while the text instruction remains fully open-vocabulary. 


\subsubsection{Trajectory Parameterization}\label{sec:traj_parameterize} 


Given the structural similarities between humans and humanoid robots, we adopt the trajectory parameterization commonly used in human motion generation~\cite{humanml}. 
Specifically, we utilize the canonicalized SMPL representation~\cite{humanml}, which defines $22$ keypoints~\cite{SMPL} within a local, root-centered coordinate frame. 
By selecting $22$ semantically corresponding joints on the $G1$ humanoid and processing them through the identical pipeline, we achieve a $G1$ trajectory representation equivalent to the SMPL format. 
This alignment allows for the direct adaptation and fine-tuning of prior human motion models for humanoid control.


After retargeting, each $G1$ trajectory is represented as a sequence of poses $\mathcal{T} = [x_{0}, x_{1}, \dots, x_{n}]$, where each pose consists of the robot joint angles and global pose: 
$x_t = (q_{\text{robot}}, T_{\text{robot}})$.
Following the pre-processing in~\cite{humanml,yang2025omniretarget,mdm}, we (1) resample all trajectories to 20 FPS, (2) crop or pad them to a maximum duration of 10 seconds, and (3) canonicalize orientations by rotating each sequence so the root at timestep $t=0$ faces the $Z+$ direction.
Each processed pose is encoded as
$ x_{t} = (\dot{r}_{a} , \dot{r}_{xz} , r_{y} , j_{p} ,j_{v} , j_{r} , c_{f} ) \in R^{263}$
where $\dot{r}{xz} \in \mathbb{R}^{2}$ is the root planar velocity,
$\dot{r}{a} \in \mathbb{R}$ is the root angular velocity,
$r_y \in \mathbb{R}$ is the root height,
$j_{p}, j_{v} \in \mathbb{R}^{3k}$ denote joint positions and velocities,
$j_{r} \in \mathbb{R}^{6k}$ is the 6D rotation representation,
and $c_f \in \mathbb{R}^{4}$ contains foot contact features, which are binary values ($2$ values per foot) indicating contact (or no-contact) with the floor near the toe and heels.
with $k = 22$ joints.

\subsubsection{Training Strategy} 
Our trajectory parameterization enables adapting existing human motion generation pipelines for humanoid control.
Here, we use the OmniControl architecture~\cite{xie2024omnicontrol} initialized with MDM~\cite{mdm} weights for training humanoid motion diffusion model. 
We next briefly summarize the diffusion model and describe how it is conditioned and trained for controllable humanoid trajectory generation.

\noindent\textbf{Diffusion Models.}
We treat motion trajectories as samples from an unknown distribution $Q(x)$ and train a denoising model $P_\theta$ to approximate it.
Following DDPM~\cite{ddpm}, we use a Markovian forward process that gradually perturbs a clean trajectory $x^{1:N}$ with Gaussian noise according to a variance schedule $\alpha_t$, and train $P_\theta$ to invert this process.
Consistent with prior work in human motion generation~\cite{xie2024omnicontrol,mdm}, the model predicts the clean signal $\hat{x}_0^{1:N}$ rather than the noise $\epsilon_t$, which has previously shown to empirically stabilize training and improve sample quality.

\noindent\textbf{Controllable Humanoid Motion Generation.}  
Given the spatio-temporal signal $\mathbf{c} \in R^{N \times J \times 3}$, and an open-vocabulary text instruction, the model generates trajectories consistent with both modalities.
We adopt the realism-guidance module from OmniControl~\cite{xie2024omnicontrol},
which is similar to ControlNet~\cite{controlnet}. 
The realism guidance block is a trainable copy of the transformer encoder with weights initialized with zeros, so that they have no controlling effect at the beginning. 
The module takes the text input $\mathbf{p}$ and spatial constrains $\mathbf{c}$ and modulates the transformer features. 
Through training the realism guidance transformer learns spatial constraints  and adds corrections to encourage physical plausibility and reduce artifacts such as foot-skating.

\subsubsection{Sampling}
OmniControl~\cite{xie2024omnicontrol} employs spatial guidance (Eq.~\ref{eq:spatial_guidance}) during sampling to satisfy spatial constraints. 
At each denoising step, forward kinematics maps the trajectory $\mu_t$ to keypoint space. 
An MSE loss $\mathcal{G}(\mu_t, \mathbf{c})$ against the target positions $\mathbf{c}$ then shifts $\mu_t$ toward these desired constraints.
\begin{equation}\label{eq:spatial_guidance}
\mu_{t} = \mu_{t} - \tau \nabla_{\mu} \mathcal{G}(\mu_{t}, \mathbf{c}).
\vspace{-2mm}
\end{equation} 
Given that we parameterize humanoid trajectories using the same representation as human motion, we can directly adopt OmniControl’s spatial guidance procedure together with CFG~\cite{cfg} sampling for humanoid trajectory generation. 

\subsubsection{Post-Processing and Filtering} 

The generated trajectories ($\mathbf{x}_t \in \mathbb{R}^{N \times 263}$) are converted into executable robot motions via a post-optimization module $\mathcal{F}_{post}$, which recovers joint angles $q_t \in \mathbb{R}^{27}$ and global poses $T_t \in \mathtt{SE(3)}$:  $\tau_{t} : q_t,, T_t = \mathcal{F}_{post}(\mathbf{x}_t)$.

We apply automated filtering and refinement steps to ensure physical plausibility and task feasibility.
Our filtering uses simple, error-based heuristics to remove noisy trajectories from the generated set.
For example, in the \texttt{Drawer Open} task, we measure the $L_2$ distance between the drawer-handle position and the robot hand link; if this error exceeds a threshold, we discard the trajectory.
Similarly, we apply task-specific conditions to remove trajectories with unsafe root heights (e.g., the root falling below a threshold), enforce foot stability to prevent slipping or floating artifacts, impose obstacle-avoidance constraints, and enforce wrist-orientation constraints when grasping is required.

\subsection{RL for Physics-Based Motion Imitation}
The diffusion model produces semantically meaningful but physics-free whole-body trajectories.  
To execute them on the humanoid, we train a physics-based controller using RL following the DreamControl framework~\cite{kalaria2025dreamcontrol}.  

\subsubsection{Reference Trajectory Parameterization for RL}
The RL policy receives a set of reference trajectories $\{\hat{\tau} \}_{i=0}^{\kappa}$, where 
$\kappa$ is the number of sampled trajectories.
Each reference trajectory is parameterized as $\hat{\tau} = \{ \tau,\, s_t^{\text{left}},\, s_t^{\text{right}} \}$, where $\tau$ is the post-processed diffusion output and $s_t^{\text{left}}, s_t^{\text{right}} \in \{0,1\}$ denote binary hand open/closed states inferred from the spatio-temporal conditioning signal $\mathbf{c}$.

\subsubsection{Action Space}  
A $27$-DOF $G1$ (pitch and roll are locked at the waist) equipped with two Inspire hands.
Action at timestep $t$ includes $q_t \in \mathbb{R}^{27}$, which denotes the target joint angles and the two scalars represent binary hand commands.  

\subsubsection{Observation Space}  
Following prior work~\cite{kalaria2025dreamcontrol,weng2025hdmi,yang2025omniretarget}, observations consist of real-robot proprioceptive signals, with object poses expressed in the robot’s local root frame to facilitate sim-to-real transfer.

\subsection{Sim2Real Deployment}
Real-world experiments are conducted on a Unitree-$G1$ robot, with the waist configured to allow only yaw rotation while roll and pitch are locked. 
Inspire hands are controlled through binary open/close commands.  
The robot also uses its on-board IMU to provide root angular velocities and the gravity vector for state estimation. 
The \shortname skill library consists of $8$ skills shown in Table~\ref{tab:task_performance}, and the deployment results are provided in the \textit{accompanying video}, and is shown in Fig.~\ref{teaser}.

\section{Experiments \& Analysis} \label{sec:exp}

\begin{table*}[t]
\centering
\caption{\footnotesize{\textbf{Comparison Across Training Data Mixtures.} We evaluate diffusion models trained on three data configurations: AMASS, AMASS + Nymeria, and Full-Mix. For downstream control, we additionally report the average RL policy success ratio (in simulation). Refer to Sec.~\ref{sec:baseline} for details related to baselines, and note that zero-shot refers to fully automated DreamControl~\cite{xie2024omnicontrol} for RL evaluation, and diffusion is evaluated using OmniControl~\cite{xie2024omnicontrol}.  All results are evaluated on the \textit{Full-Mix} test set.
}}
\label{tab:skill_accq}

\begin{tabular}{l c c c c c c c c c c}
\toprule
   \multirow{2.5}{*}{Train Data}
  & \multirow{2.5}{*}{\# Samples (K)}
  & \multirow{2.5}{*}{FID$\downarrow$}
  & \multicolumn{3}{c}{R-Precision$\uparrow$}
  & \multirow{2.5}{*}{Diversity$\rightarrow$}
  & \multirow{2.5}{*}{Control L2$\downarrow$}
  & \multirow{2.5}{*}{Skating Ratio$\downarrow$}
  & \multicolumn{2}{c}{\shortstack{Downstream RL Policy$\uparrow$}}\\
\cmidrule(lr){4-6}
\cmidrule(lr){10-11}
  & 
  &
  & Top-1 & Top-2 & Top-3
  &  (3.362)
  &
  & 
  & \textit{Group-1} & \textit{Group-2 }\\
\midrule
Zero-Shot  & - & 0.981 & 0.105 & 0.260 & 0.329 & 3.647 & 0.112 & 0.101 & 0.040 & \textbf{0.995} \\
\midrule
AMASS & 21.8 & 0.305 & 0.110 & 0.184 & 0.266 & 3.174 & 0.104 & 0.142 & 0.030 &  0.985  \\
AMASS + Nymeria & 26.4 & 0.298 & 0.101 & 0.204 & 0.281 & 3.084 & 0.108 & 0.132 & 0.860 &   0.985 \\
\rowcolor{blue!10}
\textbf{Full-Mix}  & \textbf{29.6} & \textbf{0.265} & \textbf{0.114}& \textbf{0.217}& \textbf{0.388}& \textbf{3.252} & \textbf{0.066} & \textbf{0.096} & \textbf{0.925} & 0.985  \\
\arrayrulecolor{black}\bottomrule
\end{tabular}
\end{table*}


\begin{table}[h]
\centering
\caption{\footnotesize{\textbf{Comparison Across Test Splits.} We report FID, R-Precision (Top-3), Control L2, and Skating Ratio on two distinct evaluation sets: (1) the \textit{AMASS Test Set}, and (2) a \textit{Non-AMASS Test Set} constructed from Nymeria (subset), GRAB, and OmniRetarget samples within the Full-Mix distribution.}}
\begin{tabular}{lcccc}

\toprule
Train Data & FID$\downarrow$ & R-Precision$\uparrow$  & Control L2$\downarrow$ & Skating Ratio$\downarrow$ \\
\midrule
\rowcolor{gray!15}
\multicolumn{5}{c}{\textit{AMASS Test Set}} \\
\midrule
AMASS & 0.328 & 0.291 & 0.070 & 0.100  \\
Full-Mix & 0.320  &  0.280 & 0.050 & 0.120 \\
\midrule 
\rowcolor{gray!15}
\multicolumn{5}{c}{\textit{Non-AMASS Test Set}} \\
\midrule
AMASS & 2.720 & 0.117  & 0.065  & 0.539   \\
Full-Mix & 0.316 & 0.264 & 0.042 & 0.032   \\
\bottomrule
\vspace{-6mm}
\label{tab:eval-set}
\end{tabular}
\end{table}

To analyze the impact of the \shortname's pre-retargetting recipe, we pose three key questions:
\begin{enumerate*}[label=(\textbf{Q\arabic*})] 
    \item Does a trained diffusion model operating in robot-space provide high-quality reference trajectories? (Sec~\ref{exp:ans-q1})
    \item Does scaling the diffusion training data and the number of samples drawn during RL yield any benefits (Sec~\ref{exp:task_specific})?
    \item How do the design choices of \shortname impact the system-level simplicity and overall performance (Sec.~\ref{sec:perf_analysis})?
\end{enumerate*}

\subsection{Evaluation Setup} \label{exp:setup} 
We evaluate \shortname across multiple data configurations, and tasks to analyze its fidelity, generalization, and impact on downstream control.
Here, we summarize the training data mixtures, evaluation metrics, and key implementation details used in our experiments.

\noindent\textbf{Training Data Mixtures.}
We train three diffusion model variants with progressively richer motion datasets:
\begin{itemize}[leftmargin=0.3cm]
\item AMASS: Locomotion and simple pick/place motions, identical to the corpus used by OmniControl~\cite{xie2024omnicontrol}.
\item AMASS + Nymeria (subset): Adds the subset of Nymeria containing articulated-object interactions (e.g., drawers, cabinets), following the filtering procedure in Sec.~\ref{sec:nymeria_process}.
\item Full-Mix: Augments the above with GRAB~\cite{taheri2020grab} and robot trajectories from OmniRetarget~\cite{yang2025omniretarget}.
\end{itemize}

\noindent\textbf{Baseline.} \label{sec:baseline}
We evaluate \shortname against the zero-shot OmniControl~\cite{xie2024omnicontrol} model (this model operates in human space and is trained on AMASS~\cite{AMASS,humanml} and the default model used by DreamControl~\cite{kalaria2025dreamcontrol}).
The tasks used for RL evaluations are divided into two groups. \textit{Group-1} consists of \texttt{Drawer Open} and \texttt{Deep Squat \& Pick}, and \textit{Group-2} consists of \texttt{Pick} and \texttt{Punch}. 
Tasks in \textit{Group-1} are not within the original distribution of AMASS~\cite{AMASS,humanml} for which DreamControl~\cite{kalaria2025dreamcontrol} leveraged task-specific IK and manual filtering.
Tasks in \textit{Group-2} are those that are available within the AMASS~\cite{AMASS,humanml}. 
Trial-and-error-based prompting to obtain suitable reference trajectories and the generated trajectories from zero-shot OmniControl~\cite{xie2024omnicontrol} are  passed through our automated filtering procedure and used to train the  a downstream RL baseline.  
This baseline is the fully automated variant of  DreamControl~\cite{kalaria2025dreamcontrol} (by replacing the manual filtering used in the original paper). 


\noindent\textbf{Task Library.}
After training the diffusion prior, we train a skill library consisting of 8 tasks, including articulated-object interaction, single-hand pick-and-place, and bimanual manipulation skills.
Each RL evaluation is conducted over 100 randomized simulation environments per task.
Unless stated otherwise, all RL policy results across tasks are derived from the Full-Mix model.

\noindent\textbf{Evaluation Metrics.} 
We evaluate the diffusion model using standard metrics from the human-motion generation literature~\cite{humanml,harithas2025motionglot,xie2024omnicontrol,pinyoanuntapong2024controlmm}.

\begin{itemize}[leftmargin=0.3cm]
\item FID (\(\downarrow\))~\cite{FID} 
measures the distributional similarity between generated trajectories and real motion data.
We compute FID on trajectories generated directly for the $G1$ humanoid, using a ground-truth evaluation set obtained by retargeting human motion data to the $G1$ form factor.
Because our trajectory parameterization follows~\cite{humanml}, we directly apply the HumanML3D feature extractor~\cite{humanml} to compute FID.

\item R-Precision (\(\uparrow\))
evaluates how well a generated trajectory matches its input text instruction.
For each generated sample $\hat{y}$, we retrieve the correct text description from a set of 32 candidates (1 ground-truth and 31 distractors) by ranking feature distances. 
Higher top-$k$ retrieval accuracy indicates better text–motion alignment.

\item Diversity (\(\rightarrow\))
measures the average distance between features of $N=32$ randomly sampled generated trajectories, (\(\rightarrow\)) indicates the value closer to groundtruth is better.
\item Control L2 (\(\downarrow\))
measures spatial accuracy between generated trajectories and input control signal.
\item Skating Ratio (\(\downarrow\)) 
is a proxy for physical plausibility, measuring undesired foot sliding.
\end{itemize}

\noindent For the downstream RL policy, we report two metrics:

\begin{itemize}[leftmargin=0.3cm]
\item Success Ratio (\(\uparrow\)) is the fraction of rollouts in which the humanoid successfully completes the task.
\item Control Error (\(\downarrow\))   Is the $L_{2}$ distance between the expected point of interaction and the actual point of interaction when rolling out the policy. 
\end{itemize}

\noindent\textbf{Implementation Details.}  
The large-scale human-to-robot dataset by retargeting human motion sequences across 5 NVIDIA H100 GPUs in parallel, cropping or padding each trajectory to $10$ seconds.
The diffusion model is trained on a single NVIDIA H100 GPU using the AdamW optimizer~\cite{diederik2014adam} with a learning rate of $1\times10^{-5}$.
The RL controller is trained in IsaacLab~\cite{mittal2025isaaclab}, which runs on IsaacSim.
We use PPO~\cite{ppo} on a single NVIDIA A6000 GPU for $2{,}000$–$5{,}000$ iterations depending on the task, on approximately $8,912$ parallel simulation environments.


\subsection{Scaling Data Improves Diffusion Models} \label{exp:ans-q1} 
To understand how expanding the training corpus affects the fidelity and breadth of skills learned by \shortname, we evaluate three data-mixture variants from Sec.~\ref{exp:setup}: AMASS, AMASS + Nymeria, and Full-Mix.
Table~\ref{tab:skill_accq} shows that enlarging the training distribution consistently improves trajectory quality.
Specifically, FID decreases across all mixtures, indicating clear gains in generated motion fidelity.
Compared to the zero-shot OmniControl baseline, all trained variants demonstrate substantial improvements.
This highlights the value of our pipeline, which transfers large-scale human motion data into robot space and trains a dedicated humanoid motion diffusion model, rather than relying on human-domain generation.
Among all variants, the Full-Mix model delivers the strongest and most stable performance across metrics, demonstrating that broader human–robot motion coverage yields higher-fidelity trajectories.

To assess whether expanding the training corpus preserves previously learned behaviors while enabling new ones, we evaluate the AMASS and Full-Mix models on two test splits (Table~\ref{tab:eval-set}). 
The first test split, the \textit{AMASS test set}, contains only AMASS~\cite{humanml} test sequences. 
The second split, the \textit{Non-AMASS test set}, includes test data from Nymeria~\cite{ma2024nymeria}, GRAB~\cite{taheri2020grab}, and OmniRetarget~\cite{yang2025omniretarget}. 
On the AMASS test set, the Full-Mix model performs on par with the AMASS-only variant, indicating that incorporating additional datasets does not compromise performance on the original domain. 
This suggests that our normalization pipeline (Sec.~\ref{sec:nymeria_process}) effectively aligns heterogeneous datasets, allowing new datasets to integrate smoothly without disrupting existing capabilities. 
Moreover, on the non-AMASS test set, the Full-Mix model improves over the AMASS-only variant.



\subsection{Improved Diffusion Models Yield Stronger RL Policies} \label{exp:task_specific}

\subsubsection{\textbf{Skill Acquisition from Diffusion}} To study skill acquisition We benchmark two representative task groups (\textit{Group-1/2} refer to baselines in Sec.~\ref{sec:exp}) and report the RL success ratio for each.
\textit{Group-1} includes relatively more complex actions, requiring whole body coordination such as squatting, grasping, and lifting (\texttt{Deep Squat \& lift}) or interaction with an articulated object (\texttt{Drawer Open}).
Because these tasks do not appear in AMASS, the AMASS-only model and the zero-shot model (fully automated original DreamControl~\cite{kalaria2025dreamcontrol}) is limited by its training distribution, thereby leading to low results, while incorporating interaction-rich
Data from multiple sources allows the policy to succeed.
Policies trained using trajectories from increasingly diverse diffusion models exhibit steadily higher success ratios (see Table~\ref{tab:skill_accq}).  
Overall, \shortname matches the performance of DreamControl~\cite{kalaria2025dreamcontrol} on simpler \textit{Group-2} tasks, while handling the complex, interaction-heavy \textit{Group-1} tasks far more effectively—all through a fully automated pipeline.

\renewcommand{\thefigure}{3}
\begin{figure}[t]
    \centering
    \includegraphics[
        width=1.0\columnwidth
    ]{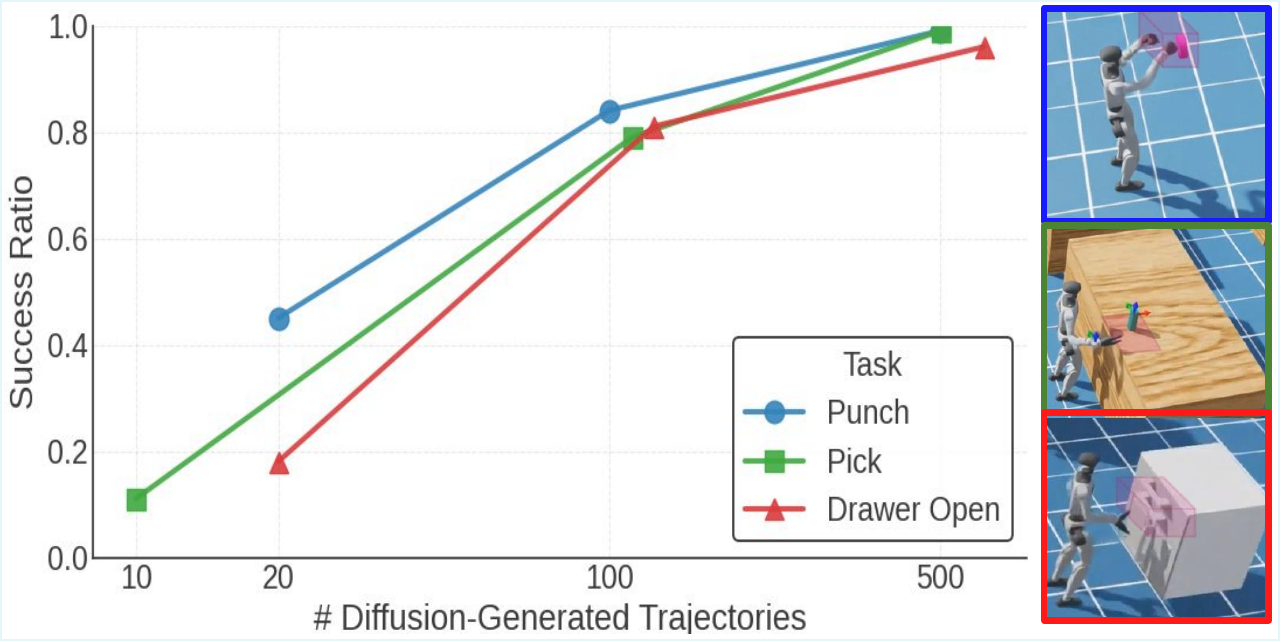}
    \vspace{-0.1in}
    \caption{ \footnotesize{ \textbf{Effect of Trajectory Scaling on RL Performance.}
    More diffusion-generated trajectories consistently lead to higher policy success ratios across 3 different tasks. The figure on the right indicates the variance of test distribution on multiple tasks. }}
    \label{fig:traj_scale}
    \vspace{-0.7cm}
\end{figure}


\begin{table}[h]
\centering
\caption{\footnotesize{\textbf{Task-wise Performance}. We report the Success Ratio and Control Error (m) averaged over 100 randomized simulation environments for all 8 tasks in our skill library.
}}
\begin{tabular}{lccc}
\toprule
Task & Success Ratio $\uparrow$ & Control Error $\downarrow$ \\
\midrule
\texttt{Pick} & 0.99 &  0.03 \\
\texttt{Punch} &  0.99  & 0.04 \\   
\texttt{Deep Squat \& Pick} & 0.94 & 0.06 \\  
\texttt{Drawer Open} &  0.96  & 0.02 \\ 
\texttt{Pour} &  0.92  & 0.03 \\  
\texttt{Vertical Wipe} & 1.00  & 0.04 \\ 
\texttt{Deep Squat} &  0.99  & 0.08 \\  
\texttt{Step \& Punch}  & 0.91  & 0.06 \\   
\bottomrule
\end{tabular}
\label{tab:task_performance}
\vspace{-2mm}
\end{table}

\subsubsection{\textbf{Scaling Diffusion Trajectories Yield Stronger RL Policies}} \label{exp:scaling}
We study how scaling the number of diffusion-generated trajectories affects generalization in downstream RL. 
Using the Full-Mix model, we generate varying numbers of trajectories and train RL policies for tasks with objects placed at diverse spatial locations.
As shown in Fig.~\ref{fig:traj_scale}, scaling of sampled trajectories consistently improves policy success, eventually approaching optimal saturation. 
The model has been evaluated with target locations sampled with an average variance of $x \sim 0.15m,\ y \sim 0.2m$ and $z \sim 0.08m$.





\subsection{Performance Analysis}\label{sec:perf_analysis}

\subsubsection{Overall Performance in Simulation} 
Table~\ref{tab:task_performance} reports task-wise performance across the 8 skills in our library.
Success criteria are defined per task; for example, in \texttt{Drawer Open}, the drawer must move at least 5\,cm.
Across all skills, \shortname achieves consistently high success rates with low control error, demonstrating robust and reliable execution of diverse loco-manipulation tasks.


\subsubsection{Inference-Time Prompt Calibration vs. Pre-retargeting}\label{exp:prompt_calib}
We consider an inference-time prompt calibration baseline that keeps the human-motion diffusion model fixed and instead tunes spatial prompting parameters in human space. 
We sample a human trajectory $x(t)$ from OmniControl~\cite{xie2024omnicontrol}, generate $N$ candidates by perturbing the prompt scale $\{x^{1}(t), \ldots, x^{N}(t)\}$, retarget each candidate to robot space, and select the one that best matches the target spatial constraints.
Table~\ref{tab:overall_prmpt} reveals two key limitations. 
First, because spatial prompts constrain multiple joints together, uniform scaling often produces off-balance motions or misses the intended constraints, resulting in higher FID and larger control errors (Ctrl L2). 
Second, this search is sample-inefficient: most candidates are rejected by the automated filter as indicated by Valid Traj.($\%$).
On the AMASS test set, the \texttt{Pick} is used for evaluation and \texttt{Drawer Open} is used for the Full-Mix evaluation; sample efficiency is reported over $100$ samples. 
This indicates that the search space is high and requires careful parameter selection and pre-retargetting is a simpler design choice.
The qualitative results are given in Fig.~\ref{fig:qualitative_comparision}, and we notice that in both tasks, the prompt calibration method leads to inaccurate samples with unstable poses, which are not near the object. 
Together with Table~\ref{tab:skill_accq}, these results show that pre-retargeting eliminates trial-and-error prompting and improves fidelity and RL performance.

\renewcommand{\thefigure}{4}
\begin{figure}[t]
    \centering
    \includegraphics[
        width=\columnwidth
    ]{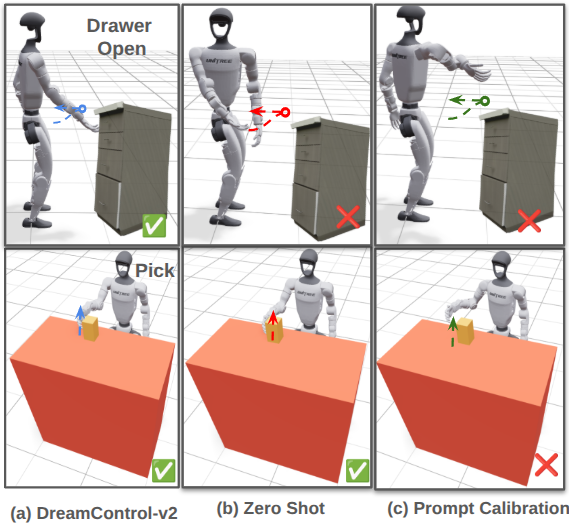}
    \vspace{-0.15in}
    \caption{ \footnotesize{ \textbf{Spatial Prompting Qualitative Results.} 
    The figure depicts successful (\twemoji{check mark button}) and rejected samples (\twemoji{cross mark}) obtained from various spatial prompting techniques. 
    The \shortname model is directly prompted in robot space (blue line), trial-and-error is used for the zero-shot model (red line), and prompt calibration is indicated with green lines. 
    }}
    \label{fig:qualitative_comparision}
    \vspace{-0.1cm}
\end{figure}

\begin{table}[t!]
\centering
\caption{\footnotesize{\textbf{Prompt calibration vs. pre-retargeting.} 
Prompt-Calib.\ uses samples from the zero-shot OmniControl model~\cite{xie2024omnicontrol}. FID and Ctrl L2 are evaluated on the respective AMASS and Full-Mix test sets. Valid Traj. (\%) is reported for the \texttt{Pick} task corresponding to AMASS and the \texttt{Drawer Open} task for Full-Mix. Our Pre-retargeting method is evaluated on Full-Mix and Valid Traj.(\%) represents the average across both tasks.}}
\begin{tabular}{lcccc}
\toprule
Method 
& FID$\downarrow$
& Ctrl L2$\downarrow$ 
 & Valid Traj.(\%) \\
\midrule
\footnotesize{Prompt-Calib. (\textit{AMASS})} & 2.010 & 0.180  & 8.0 \\
\footnotesize{Prompt-Calib. (\textit{Full-Mix})} & 3.260 & 0.260  & 4.0 \\
\rowcolor{blue!10}
\footnotesize{Pre-retargetting (\textit{Full-Mix})}  & \textbf{0.265} & \textbf{0.066} & \textbf{68.0}\\
\bottomrule
\label{tab:overall_prmpt}
\vspace{-7mm}
\end{tabular}
\end{table}

\subsubsection{Network Initialization} 
Our trajectory parameterization for the humanoid robot (given in Sec.~\ref {sec:traj_parameterize}) allows us to use the strong priors from human motion diffusion model (MDM~\cite{mdm}) for humanoid motion generation. 
Table.~\ref{tab:network-init} reports an ablation comparing models trained from scratch versus those fine-tuned from the MDM checkpoint. 
Fine-tuning yields clear gains in fidelity and physical plausibility, suggesting that human-motion priors provide a strong initialization that adapts effectively to the robot embodiment.

\begin{table}[t!]
\centering
\caption{\footnotesize{\textbf{Network Initialization }. We report FID, R-Precision (Top-3), Control L2 and Skating Ratio for models randomly initialized ($\text{\no}$) and with the pretrained weights from MDM~\cite{mdm} ($\text{\yes}$).}}
\begin{tabular}{cccccc}

\toprule
Pretrained & Iters $\times{10^{3}}$ 
& FID$\downarrow$
& R@3$\uparrow$
& Ctrl L2$\downarrow$ 
& Skt Ratio$\downarrow$ \\
\midrule
\no & 104 & 0.741  & 0.254 & 0.114 & 0.112  \\
\yes & 102 & \textbf{0.265} & \textbf{0.388}& \textbf{0.066}& \textbf{0.096}\\
\bottomrule
\label{tab:network-init}
\vspace{-8mm}
\end{tabular}
\end{table}

\vspace{-3mm}
\bibliographystyle{IEEEtran}
\bibliography{references}

@inproceedings{seo2023deep,
  title={Deep imitation learning for humanoid loco-manipulation through human teleoperation},
  author={Seo, Mingyo and Han, Steve and Sim, Kyutae and Bang, Seung Hyeon and Gonzalez, Carlos and Sentis, Luis and Zhu, Yuke},
  booktitle={2023 IEEE-RAS 22nd International Conference on Humanoid Robots (Humanoids)},
  pages={1--8},
  year={2023},
  organization={IEEE}
}

@article{fu2024humanplus,
  title={Humanplus: Humanoid shadowing and imitation from humans},
  author={Fu, Zipeng and Zhao, Qingqing and Wu, Qi and Wetzstein, Gordon and Finn, Chelsea},
  journal={arXiv preprint arXiv:2406.10454},
  year={2024}
}

@article{FID,
  title={Gans trained by a two time-scale update rule converge to a local nash equilibrium},
  author={Martin, Heusel and Hubert, Ramsauer and Thomas, Unterthiner and Bernhard, Nessler and Sepp, Hochreiter},
  journal={Advances in neural information processing systems},
  volume={30},
  pages={6626--6637},
  year={2017},
  publisher={NIPS}
}

@inproceedings{he2024learning,
  title={Learning human-to-humanoid real-time whole-body teleoperation},
  author={He, Tairan and Luo, Zhengyi and Xiao, Wenli and Zhang, Chong and Kitani, Kris and Liu, Changliu and Shi, Guanya},
  booktitle={2024 IEEE/RSJ International Conference on Intelligent Robots and Systems (IROS)},
  pages={8944--8951},
  year={2024},
  organization={IEEE}
}

@article{he2024omnih2o,
      title={OmniH2O: Universal and Dexterous Human-to-Humanoid Whole-Body Teleoperation and Learning},
      author={He, Tairan and Luo, Zhengyi and He, Xialin and Xiao, Wenli and Zhang, Chong and Zhang, Weinan and Kitani, Kris and Liu, Changliu and Shi, Guanya},
      journal={arXiv preprint arXiv:2406.08858},
      year={2024}
    }

@article{cheng2024express,
    title={Expressive Whole-Body Control for Humanoid Robots},
    author={Cheng, Xuxin and Ji, Yandong and Chen, Junming and Yang, Ruihan and Yang, Ge and Wang, Xiaolong},
    journal={arXiv preprint arXiv:2402.16796},
    year={2024}
}

@article{ji2024exbody2,
  title={ExBody2: Advanced Expressive Humanoid Whole-Body Control}, 
  author={Ji, Mazeyu and Peng, Xuanbin and Liu, Fangchen and Li, Jialong and Yang, Ge and Cheng, Xuxin and Wang, Xiaolong},
  journal={arXiv preprint arXiv:2412.13196},
  year={2024},
  }

@article{ze2025twist2,
  title={TWIST2: Scalable, Portable, and Holistic Humanoid Data Collection System},
  author={Ze, Yanjie and Zhao, Siheng and Wang, Weizhuo and Kanazawa, Angjoo and Duan, Rocky and Abbeel, Pieter and Shi, Guanya and Wu, Jiajun and Liu, C Karen},
  journal={arXiv preprint arXiv:2511.02832},
  year={2025}
}

@article{pi0,
  title={$\pi_0$: A Vision-Language-Action Flow Model for General Robot Control},
  author={Black, Kevin and Brown, Noah and Driess, Danny and Esmail, Adnan and Equi, Michael and Finn, Chelsea and Fusai, Niccolo and Groom, Lachy and Hausman, Karol and Ichter, Brian and others},
  journal={arXiv preprint arXiv:2410.24164},
  year={2024}
}

@article{pi05,
  title={$\pi_0.5$: a Vision-Language-Action Model with Open-World Generalization},
  author={Intelligence, Physical and Black, Kevin and Brown, Noah and Darpinian, James and Dhabalia, Karan and Driess, Danny and Esmail, Adnan and Equi, Michael and Finn, Chelsea and Fusai, Niccolo and others},
  journal={arXiv preprint arXiv:2504.16054},
  year={2025}
}

@article{diffusionpolicy,
  title={Diffusion policy: Visuomotor policy learning via action diffusion},
  author={Chi, Cheng and Xu, Zhenjia and Feng, Siyuan and Cousineau, Eric and Du, Yilun and Burchfiel, Benjamin and Tedrake, Russ and Song, Shuran},
  journal={The International Journal of Robotics Research},
  volume={44},
  number={10-11},
  pages={1684--1704},
  year={2025},
  publisher={Sage Publications Sage UK: London, England}
}

@article{pinyoanuntapong2024controlmm,
  title={Controlmm: Controllable masked motion generation},
  author={Pinyoanuntapong, Ekkasit and Saleem, Muhammad Usama and Karunratanakul, Korrawe and Wang, Pu and Xue, Hongfei and Chen, Chen and Guo, Chuan and Cao, Junli and Ren, Jian and Tulyakov, Sergey},
  journal={arXiv preprint arXiv:2410.10780},
  year={2024}
}

@article{mittal2025isaaclab,
  title={Isaac Lab: A GPU-Accelerated Simulation Framework for Multi-Modal Robot Learning},
  author={Mayank Mittal et.al },
  journal={arXiv preprint arXiv:2511.04831},
  year={2025},
  url={https://arxiv.org/abs/2511.04831}
}

@inproceedings{controlnet,
  title={Adding conditional control to text-to-image diffusion models},
  author={Zhang, Lvmin and Rao, Anyi and Agrawala, Maneesh},
  booktitle={Proceedings of the IEEE/CVF international conference on computer vision},
  pages={3836--3847},
  year={2023}
}

@inproceedings{pan2024synthesizing,
    title={Synthesizing physically plausible human motions in 3d scenes},
    author={Pan, Liang and Wang, Jingbo and Huang, Buzhen and Zhang, Junyu and Wang, Haofan and Tang, Xu and Wang, Yangang},
    booktitle={2024 International Conference on 3D Vision (3DV)},
    pages={1498--1507},
    year={2024},
    organization={IEEE}
}

@inproceedings{videomimic,
  title     = {Visual imitation enables contextual humanoid control},
  author    = {Allshire, Arthur and Choi, Hongsuk and Zhang, Junyi and McAllister, David 
               and Zhang, Anthony and Kim, Chung Min and Darrell, Trevor and Abbeel, 
               Pieter and Malik, Jitendra and Kanazawa, Angjoo},
  booktitle = {Proceedings of the Conference on Robot Learning (CoRL)},
  year      = {2025}
}

@article{tevet2024closd,
  title={Closd: Closing the loop between simulation and diffusion for multi-task character control},
  author={Tevet, Guy and Raab, Sigal and Cohan, Setareh and Reda, Daniele and Luo, Zhengyi and Peng, Xue Bin and Bermano, Amit H and van de Panne, Michiel},
  journal={arXiv preprint arXiv:2410.03441},
  year={2024}
}

@article{twist,
  title={TWIST: Teleoperated Whole-Body Imitation System},
  author={Ze, Yanjie and Chen, Zixuan and Ara{\~A}{\v{s}}jo, Jo{\~A}{\c{G}}o Pedro and Cao, Zi-ang and Peng, Xue Bin and Wu, Jiajun and Liu, C Karen},
  journal={arXiv preprint arXiv:2505.02833},
  year={2025}
}

@article{yang2025omniretarget,
  title={OmniRetarget: Interaction-Preserving Data Generation for Humanoid Whole-Body Loco-Manipulation and Scene Interaction},
  author={Yang, Lujie and Huang, Xiaoyu and Wu, Zhen and Kanazawa, Angjoo and Abbeel, Pieter and Sferrazza, Carmelo and Liu, C Karen and Duan, Rocky and Shi, Guanya},
  journal={arXiv preprint arXiv:2509.26633},
  year={2025}
}

@article{kalaria2025dreamcontrol,
  title={DreamControl: Human-Inspired Whole-Body Humanoid Control for Scene Interaction via Guided Diffusion},
  author={Kalaria, Dvij and Harithas, Sudarshan S and Katara, Pushkal and Kwak, Sangkyung and Bhagat, Sarthak and Sastry, Shankar and Sridhar, Srinath and Vemprala, Sai and Kapoor, Ashish and Huang, Jonathan Chung-Kuan},
  journal={arXiv preprint arXiv:2509.14353},
  year={2025}
}

@article{weng2025hdmi,
  title={HDMI: Learning Interactive Humanoid Whole-Body Control from Human Videos},
  author={Weng, Haoyang and Li, Yitang and Sobanbabu, Nikhil and Wang, Zihan and Luo, Zhengyi and He, Tairan and Ramanan, Deva and Shi, Guanya},
  journal={arXiv preprint arXiv:2509.16757},
  year={2025}
}

@InProceedings{humanml,
    author    = {Guo, Chuan and Zou, Shihao and Zuo, Xinxin and Wang, Sen and Ji, Wei and Li, Xingyu and Cheng, Li},
    title     = {Generating Diverse and Natural 3D Human Motions From Text},
    booktitle = {Proceedings of the IEEE/CVF Conference on Computer Vision and Pattern Recognition (CVPR)},
    month     = {June},
    year      = {2022},
    pages     = {5152-5161}
}

@conference{AMASS,
  title = {{AMASS}: Archive of Motion Capture as Surface Shapes},
  author = {Mahmood, Naureen and Ghorbani, Nima and Troje, Nikolaus F. and Pons-Moll, Gerard and Black, Michael J.},
  booktitle = {International Conference on Computer Vision},
  pages = {5442--5451},
  month = oct,
  year = {2019},
  month_numeric = {10}
}

@inproceedings{taheri2020grab,
  title={GRAB: A dataset of whole-body human grasping of objects},
  author={Taheri, Omid and Ghorbani, Nima and Black, Michael J and Tzionas, Dimitrios},
  booktitle={European conference on computer vision},
  pages={581--600},
  year={2020},
  organization={Springer}
}

@inproceedings{ma2024nymeria,
  title={Nymeria: A massive collection of multimodal egocentric daily motion in the wild},
  author={Ma, Lingni and Ye, Yuting and Hong, Fangzhou and Guzov, Vladimir and Jiang, Yifeng and Postyeni, Rowan and Pesqueira, Luis and Gamino, Alexander and Baiyya, Vijay and Kim, Hyo Jin and others},
  booktitle={European Conference on Computer Vision},
  pages={445--465},
  year={2024},
  organization={Springer}
}

@software{jax2018github,
  author = {James Bradbury and Roy Frostig and Peter Hawkins and Matthew James Johnson and Chris Leary and Dougal Maclaurin and George Necula and Adam Paszke and Jake Vander{P}las and Skye Wanderman-{M}ilne and Qiao Zhang},
  title = {{JAX}: composable transformations of {P}ython+{N}um{P}y programs},
  url = {http://github.com/jax-ml/jax},
  version = {0.3.13},
  year = {2018},
}

@inproceedings{kim2025pyroki,
  title={PyRoki: A Modular Toolkit for Robot Kinematic Optimization},
  author={Kim*, Chung Min and Yi*, Brent and Choi, Hongsuk and Ma, Yi and Goldberg, Ken and Kanazawa, Angjoo},
  booktitle={2025 IEEE/RSJ International Conference on Intelligent Robots and Systems (IROS)},
  year={2025},
  url={https://arxiv.org/abs/2505.03728},
}

@inproceedings{Luo2023PerpetualHC,
    author={Zhengyi Luo and Jinkun Cao and Alexander W. Winkler and Kris Kitani and Weipeng Xu},
    title={Perpetual Humanoid Control for Real-time Simulated Avatars},
    booktitle={International Conference on Computer Vision (ICCV)},
    year={2023}
}

@article{uhc,
  title={Dynamics-regulated kinematic policy for egocentric pose estimation},
  author={Luo, Zhengyi and Hachiuma, Ryo and Yuan, Ye and Kitani, Kris},
  journal={Advances in Neural Information Processing Systems},
  volume={34},
  pages={25019--25032},
  year={2021}
}

@incollection{SMPL,
  title={SMPL: A skinned multi-person linear model},
  author={Loper, Matthew and Mahmood, Naureen and Romero, Javier and Pons-Moll, Gerard and Black, Michael J},
  booktitle={Seminal Graphics Papers: Pushing the Boundaries, Volume 2},
  pages={851--866},
  year={2023}
}

@misc{xsens,
  title        = {Movella {Xsens} MVN Link Motion Capture System},
  author       = {{Movella Technologies B.V.}},
  year         = {2025},
  howpublished = {\url{https://www.movella.com/products/motion-capture/xsens-mvn-link}},
  note         = {Accessed: 2025-11-04}
}

@inproceedings{
xie2024omnicontrol,
title={OmniControl: Control Any Joint at Any Time for Human Motion Generation},
author={Yiming Xie and Varun Jampani and Lei Zhong and Deqing Sun and Huaizu Jiang},
booktitle={The Twelfth International Conference on Learning Representations},
year={2024},
url={https://openreview.net/forum?id=gd0lAEtWso}
}

@inproceedings{harithas2025motionglot,
  title={Motionglot: A multi-embodied motion generation model},
  author={Harithas, Sudarshan and Sridhar, Srinath},
  booktitle={2025 IEEE International Conference on Robotics and Automation (ICRA)},
  pages={16109--16116},
  year={2025},
  organization={IEEE}
}

@article{li2023object,
  title={Object Motion Guided Human Motion Synthesis},
  author={Li, Jiaman and Wu, Jiajun and Liu, C Karen},
  journal={ACM Trans. Graph.},
  volume={42},
  number={6},
  year={2023}
}

@article{mdm,
  title={Human motion diffusion model},
  author={Tevet, Guy and Raab, Sigal and Gordon, Brian and Shafir, Yonatan and Cohen-Or, Daniel and Bermano, Amit H},
  journal={arXiv preprint arXiv:2209.14916},
  year={2022}
}

@inproceedings{ddpm,
  title={High-resolution image synthesis with latent diffusion models},
  author={Rombach, Robin and Blattmann, Andreas and Lorenz, Dominik and Esser, Patrick and Ommer, Bj{\"o}rn},
  booktitle={Proceedings of the IEEE/CVF conference on computer vision and pattern recognition},
  pages={10684--10695},
  year={2022}
}

@article{cfg,
  title={Classifier-free diffusion guidance},
  author={Ho, Jonathan and Salimans, Tim},
  journal={arXiv preprint arXiv:2207.12598},
  year={2022}
}

@article{luo2024omnigrasp,
  title={Omnigrasp: Grasping diverse objects with simulated humanoids},
  author={Luo, Zhengyi and Cao, Jinkun and Christen, Sammy and Winkler, Alexander and Kitani, Kris and Xu, Weipeng},
  journal={Advances in Neural Information Processing Systems},
  volume={37},
  pages={2161--2184},
  year={2024}
}

@article{gpt3,
  title={Language models are few-shot learners},
  author={Brown, Tom B},
  journal={arXiv preprint arXiv:2005.14165},
  year={2020}
}

@inproceedings{stable-diff,
  title={High-resolution image synthesis with latent diffusion models},
  author={Rombach, Robin and Blattmann, Andreas and Lorenz, Dominik and Esser, Patrick and Ommer, Bj{\"o}rn},
  booktitle={Proceedings of the IEEE/CVF conference on computer vision and pattern recognition},
  pages={10684--10695},
  year={2022}
}

@article{kim2024openvla,
  title={Openvla: An open-source vision-language-action model},
  author={Kim, Moo Jin and Pertsch, Karl and Karamcheti, Siddharth and Xiao, Ted and Balakrishna, Ashwin and Nair, Suraj and Rafailov, Rafael and Foster, Ethan and Lam, Grace and Sanketi, Pannag and others},
  journal={arXiv preprint arXiv:2406.09246},
  year={2024}
}

@article{humanoid-x,
  title={Learning from massive human videos for universal humanoid pose control},
  author={Mao, Jiageng and Zhao, Siheng and Song, Siqi and Shi, Tianheng and Ye, Junjie and Zhang, Mingtong and Geng, Haoran and Malik, Jitendra and Guizilini, Vitor and Wang, Yue},
  journal={arXiv preprint arXiv:2412.14172},
  year={2024}
}

@article{t2mgpt,
  title={T2m-gpt: Generating human motion from textual descriptions with discrete representations},
  author={Zhang, Jianrong and Zhang, Yangsong and Cun, Xiaodong and Huang, Shaoli and Zhang, Yong and Zhao, Hongwei and Lu, Hongtao and Shen, Xi},
  journal={arXiv preprint arXiv:2301.06052},
  year={2023}
}

@inproceedings{Pinyoanuntapong2025MaskControl,
  title     = {MaskControl: Spatio-Temporal Control for Masked Motion Synthesis},
  author    = {Pinyoanuntapong, Ekkasit and Saleem, Muhammad and Karunratanakul, Korrawe and Wang, Pu and Xue, Hongfei and Chen, Chen and Guo, Chuan and Cao, Junli and Ren, Jian and Tulyakov, Sergey},
  booktitle = {Proceedings of the IEEE/CVF International Conference on Computer Vision (ICCV)},
  pages     = {9955--9965},
  year      = {2025}
}

@article{motiongpt,
    title={MotionGPT: Human Motion as a Foreign Language},
    author={Jiang, Biao and Chen, Xin and Liu, Wen and Yu, Jingyi and Yu, Gang and Chen, Tao},
    journal={Advances in Neural Information Processing Systems},
    volume={36},
    year={2024}
}

@article{luo2025sonic,
    title={SONIC: Supersizing Motion Tracking for Natural Humanoid Whole-Body Control},
    author={Luo, Zhengyi and Yuan, Ye and Wang, Tingwu and Li, Chenran and Chen, Sirui and Casta\~neda, Fernando and Cao, Zi-Ang and Li, Jiefeng and Minor, David and Ben, Qingwei and Da, Xingye and Ding, Runyu and Hogg, Cyrus and Song, Lina and Lim, Edy and Jeong, Eugene and He, Tairan and Xue, Haoru and Xiao, Wenli and Wang, Zi and Yuen, Simon and Kautz, Jan and Chang, Yan and Iqbal, Umar and Fan, Linxi and Zhu, Yuke},
    journal={arXiv preprint arXiv:2511.07820},
    year={2025}
}

@article{diederik2014adam,
  title={Adam: A method for stochastic optimization},
  author={Diederik, P Kingma},
  journal={(No Title)},
  year={2014}
}

@article{ppo,
  title={Proximal policy optimization algorithms},
  author={Schulman, John and Wolski, Filip and Dhariwal, Prafulla and Radford, Alec and Klimov, Oleg},
  journal={arXiv preprint arXiv:1707.06347},
  year={2017}
}

@article{al2012trajectory,
  title={Trajectory optimization for full-body movements with complex contacts},
  author={Al Borno, Mazen and De Lasa, Martin and Hertzmann, Aaron},
  journal={IEEE transactions on visualization and computer graphics},
  volume={19},
  number={8},
  pages={1405--1414},
  year={2012},
  publisher={IEEE}
}

@article{gpt-4,
  title={Gpt-4 technical report},
  author={Achiam, Josh and Adler, Steven and Agarwal, Sandhini and Ahmad, Lama and Akkaya, Ilge and Aleman, Florencia Leoni and Almeida, Diogo and Altenschmidt, Janko and Altman, Sam and Anadkat, Shyamal and others},
  journal={arXiv preprint arXiv:2303.08774},
  year={2023}
}

\clearpage
\section{Appendix}

\subsection{Key-point Correspondence Between Human \& G1 }

In \shortname correspondances are needed to first retarget human data to the G1 embodiment. 
Table ~\ref{tab:smpl_g1_links} show the keypoint correspondences used for retargetting, Table~\ref{tab:smpl_g1_links_traj_rep} shows the correspondences used for the trajectory representation given in  Sec.~\ref{sec:traj_parameterize}. 

\begin{table}[h]
\centering
\caption{Human and G1 link correspondances for retargetting}
\label{tab:smpl_g1_links}
\begin{tabular}{ll}
\hline
\textbf{SMPL} & \textbf{G1 link name} \\
\hline
pelvis & pelvis\_contour\_link \\
left hip & left\_hip\_pitch\_link \\
right hip & right\_hip\_pitch\_link \\
left knee & left\_knee\_link \\
right knee & right\_knee\_link \\
left ankle & left\_ankle\_roll\_link \\
right ankle & right\_ankle\_roll\_link \\
left shoulder & left\_sholuder\_roll\_link \\
right sholuder & right\_shoulder\_roll\_link \\
left elbow & left\_elbow\_link \\
right elbow & right\_elbow\_link \\
left wrist & left\_wrist\_yaw\_link \\
right wrist & right\_wrist\_yaw\_link \\
\hline
\end{tabular}
\end{table}

\begin{table}[h]
\centering
\caption{SMPL and G1 link keypoint correspondances for Trajectory represenatation}
\label{tab:smpl_g1_links_traj_rep}
\begin{tabular}{ll}
\hline
\textbf{Human} & \textbf{G1 link name} \\
\hline
pelvis & pelvis contour link \\
left hip & left hip pitch link \\
right hip & right hip pitch link \\
spine 1 & imu link \\
left knee & left knee link \\
right knee & right knee link \\
spine 2 & torso link \\
left ankle & left ankle pitch link \\
right ankle & right ankle pitch link \\
spine 3 & imu link \\
left foot & left ankle roll link \\
right foot & right ankle roll link \\
neck & mid360 link \\
left collar & left shoulder yaw link \\
right collar & right shoulder yaw link \\
head & d435 link \\
left shoulder & left shoulder pitch link \\
right shoulder & right shoulder pitch link \\
left elbow & left elbow link \\
right elbow & right elbow link \\
left wrist & left wrist yaw link \\
right wrist & right wrist yaw link \\
\hline
\end{tabular}
\end{table}

\begin{figure}[t!]
    \centering
    \includegraphics[
        width=\columnwidth
    ]{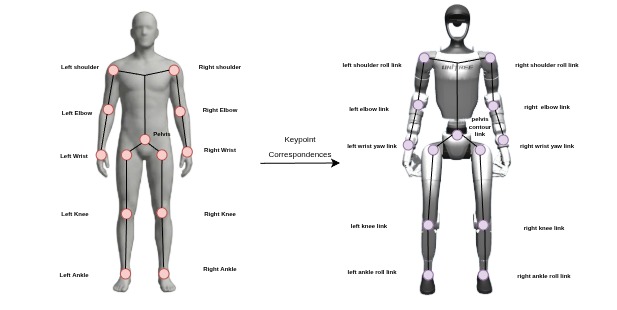}
    \vspace{-0.28in}
    \caption{Key-point correspondences used for retargetting human data to G1. A similar correspondence mapping is used for diffusion trajectory representation  Sec.~\ref{sec:traj_parameterize}.  }
    \label{fig:drawer_scaling_plot}
    \vspace{-4mm}
\end{figure}

\subsection{Diffusion Model Qualitative Results} 
\begin{figure*}[t!]
    \centering
    \includegraphics[
        width=2\columnwidth ]{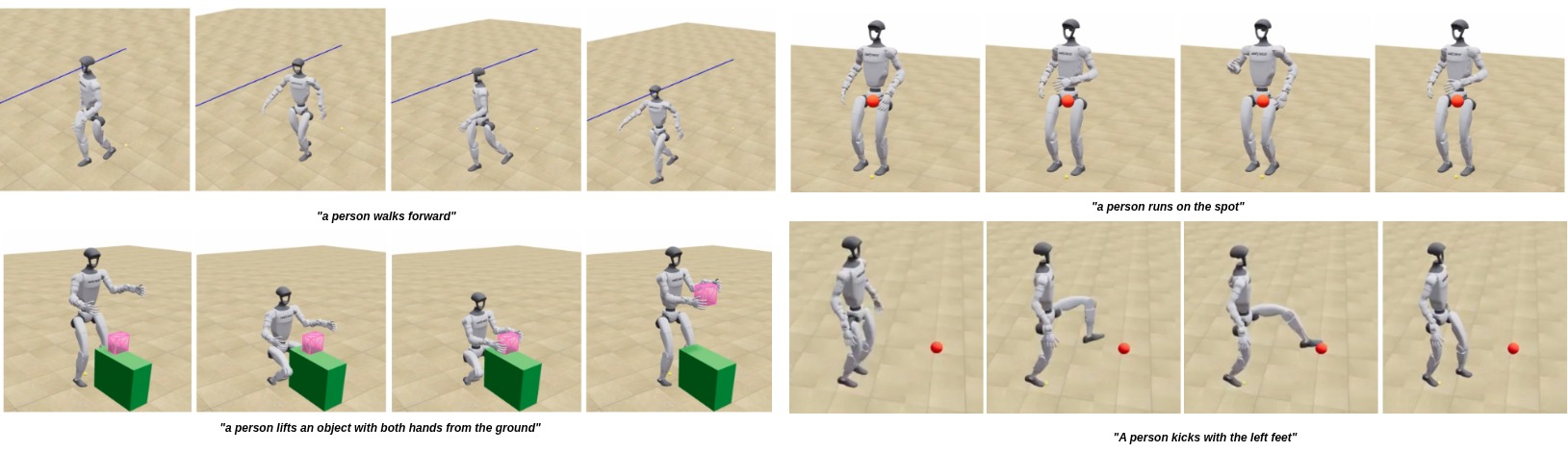}
    \caption{\textbf{Trajectory Generation Qualitative Results} }
    \label{fig:diffusion_qualitative}
\end{figure*}

The qualitative results of the reference trajectory generation using the diffusion model are shown in Fig.~\ref{fig:diffusion_qualitative}. 
The results demonstrate that the model can generate trajectories while satisfying a wide range of spatiotemporal constraints. 
For example, in the walk scenario, the generated trajectory adheres to a continuous constraint on the head joint.
Likewise, the model can be conditioned on discrete joint positions at specific timesteps to produce actions such as kick or run in place. 
Furthermore, the model supports conditioning on multiple joints simultaneously—such as both wrists—to generate coordinated behaviors like a bimanual pick action.

\subsection{Diffusion Model Prompting}

\subsubsection{Text Prompts}

The task-specific text prompts used are given in Table~\ref{tab:text-prompts}. 

\begin{table*}[t!]
\centering
\caption{Task-specific Text Prompts}
\begin{tabular}{cc}
\toprule
\textbf{Task} & \textbf{Prompt} \\
\hline 
\texttt{Pick} & "a person picks an object with their right hand" \\ 
\texttt{Bimanual Pick}  & "a person uses both hands to pick an object from the floor" \\ 
\texttt{Drawer Open} &  "a person stands and pulls a drawer open with their right hand" \\
\texttt{Pour} & "a person picks an object with the right hand and performs a pouring action with the same hand" \\ 
\texttt{Vertical Wipe} & "a person wipes a vertical surface"  \\
\texttt{Deep Squat} & "a person performs squats" \\
\texttt{Punch} & "a person punches with the right wrist" \\
\texttt{Step \& Punch} & "a person steps forward and punches with the right wrist"\\
 \bottomrule
\vspace{-4mm}
\end{tabular}
\label{tab:text-prompts}
\end{table*}

\subsubsection{SpatioTemporal Prompts}

In this section we detail the spatiotemporal prompts used to sample reference trajectories from the trained diffusion model. 
Note that the \shortname diffusion model can be directly prompted in the robot space, simplifying the prompting process previously used in~\cite{kalaria2025dreamcontrol}. 

\begin{itemize}
    \item \texttt{Pick:} The right wrist is used to perform the pick task, where a cube object is placed on a table. 
    The spatial control for the right wrist is given as ($\mathbf{p_{right\_wrist}}$) is sampled uniformly in the distribution $ x \sim \mathcal{U}(  -0.22 , 0.12 ) , y \sim \mathcal{U}( 0.2, 0.4 ) , z \sim 0.94 $. 
    The time of contact is chosen as $t \sim \mathcal{U}(40, 50)$. 
    Similar to \cite{kalaria2025dreamcontrol}, to encourage side-grasp of the cup, we further condition the elbow joints using the equation $p_{elbow} = [ x - \alpha_{l} l \cos(\theta) , y, z - l \sin(\theta)]$. 
    Where $l$ is the link length from elbow to wrist and $\theta = \pi/4$.  
    To encourage a stationary position during the pick task, we further constrain both the knee positions. 
    \item \texttt{Open Drawer:} The right wrist is used to perform the drawer open task. 
    To encourage drawer opening samples with a stationary stance, we further add constraints on the knees. 
    $\mathbf{p_{right\_wrist}} x \sim \mathcal{U}(0.3, 0.50) , y \sim \mathcal{U}(-0.20,-0.30), z \sim \mathcal{U}(0.75,0.82)$. 
    $\mathbf{p}_{right\_knee} y \sim \mathcal{N}( -0.13 , 0.01) , z \sim \mathcal{N}(0.35, 0.01), x \sim \mathcal{N}(-0.02 , 0.01)$. 
    A target timestep $\mathbf{t_{g}} \sim \mathcal{U}(35,40)$ is selected. 
    The spatiotemporal prompt ($\boldsymbol{\tau}_{\text{right\_wrist}}$) of total time $T$  is given as 
    \[
    \boldsymbol{\tau}_{\text{right\_wrist}} =
    \begin{cases}
    \mathbf{p}_{\text{init}}
    + \left( \mathbf{p}_{\text{right\_wrist}} - \mathbf{p}_{\text{init}} \right)
    \frac{t}{t_g},
    & t \le t_g
    \\[6pt]
    \mathbf{p}_{\text{right}}
    + \mathbf{p}_{\text{offset}} \frac{t}{T},
    & t_g < t \le T
    \end{cases}
\]

The parameter $\mathbf{p}_{offset}$ is set to indicate the distance to open the drawer is set to $0.2m$.

    \item  \texttt{Pour:}The prompt procedure for the pour task is similar to that of the pick task. 
    The spatial control of the right wrist, denoted by $\mathbf{p}_{\text{right\_wrist}}$, 
    is sampled from a uniform distribution with 
    $x \sim \mathcal{U}(-0.10, 0.10)$, 
    $y \sim \mathcal{U}(0.2, 0.4)$, 
    and $z = 0.94$. 
    The contact time is sampled as $t \sim \mathcal{U}(40, 42)$.

    \item \texttt{Vertical Wipe:} the right wrist is prompted using $\boldsymbol{\tau}_{\text{right\_wrist}}$, setting the spatial constrain to $0$ would disable spatial constrains until the target time $t_{g}$. 
    A linear trajectory is used to 
    \[
    \boldsymbol{\tau}_{\text{right\_wrist}} =
    \begin{cases}
    0    & t \le t_g
    \\[6pt]
    \mathbf{p}_{\text{start}}
    -  \mathbf{p}_{\text{offset}} \frac{t}{T},
    & t_g < t \le T
    \end{cases}
    \]
    \item \texttt{Deep Squat:} For the deep squat, we condition the pelvis position  to be $\mathbf{p}_{pelvis} $ sampled from $x \sim \mathcal{U}( -0.15, -0.05)$, $ y \sim \mathcal{U}(-0.04, 0.04)$ , $z \sim \mathcal{U}(0.35, 0.45)$. 
    \item \texttt{Punch:} The right wrist is conditioned for the punch task. 
    The target $(\mathbf{p_{right\_wrist}} )$ are sampled as $y \sim \mathcal{U}(-0.2, 0.1)$ , $z \sim \mathcal{U}(0.95, 1.1)$ , $x \sim  \mathcal{U}(0.45, 0.55)$. 
    The time step $t_{g} \sim \mathcal{U}(90, 120)$. 
    \item  \texttt{Step \& Punch:}  The right wrist is conditioned for the punch task. 
    The target $(\mathbf{p_{right\_wrist}} )$ are sampled as $y \sim \mathcal{U}(-0.2, 0.1)$ , $z \sim \mathcal{U}(0.95, 1.1)$ , $x \sim  \mathcal{U}(0.95, 1.05)$. 
    The time step $t_{g} \sim \mathcal{U}(110, 140)$. 
\end{itemize}

\subsection{Filtering Process \& Optimization} Here, we provide details on the task-specific filtering steps used and the post optimization steps before tracking the trajectory with RL. 

\begin{table*}[!htbp]
\centering
\caption{Task-specific Rewards}
\begin{tabular}{cccc}
\toprule
\textbf{Task} & \textbf{Reward (r1)}  & \textbf{Description} \\
\hline 
\texttt{Pick} & $h_{t}^{obj} > h^{thresh}$ , $t \geq t_{g}$  & $h^{thresh} =0.95$ ,$h_{t}^{obj}$ is object height \\
\texttt{Deep Squat \& Pick}  & $h_{t}^{obj} > h^{thresh}$, $t \geq t_{g}$   &$h^{thresh} =0.65$ \\ 
\texttt{Drawer Open} & $a_{t}^{obj} > h^{thresh}$, $t \geq t_{g}$ & $a_{t}^{obj} \geq 0.05$, distance traversed by drawer \\
\texttt{Pour} & $h_{t}^{obj} > h^{thresh}$, $t \geq t_{g}$  & $h^{thresh} =0.65$ \\
\texttt{Vertical Wipe} & $ ||p_{target} - p_{t} ||_{2}^{2} \leq \lambda$  &  $p_{t}$ position of wrist at timestep $t$ and $\lambda = 0.05$ \\
\texttt{Punch} & $ ||p_{target} - p_{t} ||_{2}^{2} \leq \lambda$  &  $p_{t}$ position of wrist at timestep $t$ and $\lambda = 0.05$ \\
\texttt{Step \& Punch}  &  $ ||p_{target} - p_{t} ||_{2}^{2} \leq \lambda$  &  $p_{t}$ position of wrist at timestep $t$ and $\lambda = 0.05$ \\
 \bottomrule
\vspace{-8mm}
\end{tabular}
\label{tab:task_rewards}
\end{table*}

\begin{itemize}
    \item \texttt{Pick:} To pick the object the trajectories are expected to avoid the table on which they are placed, similar to DreamControl~\cite{kalaria2025dreamcontrol} we follow an SDF based obstacle avoidance process. 
    In addition to the obstacle avoidance we have an additional cost to ensure smoothness while grasping the object and the hand remains close to the control point $p_{wrist}$. 
\begin{equation*}
\begin{split}
    q^{ref} = \arg\min \sum_{t} & || x^{eef}_{t} - x^{eef}_{t+1} ||_{2} + d(x^{eef}, \lambda) + \\
    & || \psi^{eef} - \psi^{eef}_{des} ||^{2}
\end{split}
\end{equation*}

    The costs include a smoothing term to ensure continuity, the function $d(.)$ is the \texttt{SDF} function that returns $0$ if the distance is greater than $\lambda$, and $\psi$ refers to the cost function that minimizes the distance between the actual and the desired end-effector pose for grasping.
    Lastly we apply a generic filtering step, where if the position of the wrist is expected to avoid the obstacle by at-least $\beta = 0.03$ and the the error between the wrist position and the target must be less than $0.15m$ to be selected for RL training. 
    \item \texttt{Open Drawer:} The drawer opening cost consists of the trajectory reaching cost given by. 
    \begin{equation*}
    \begin{split}
        q^{ref} = \arg\min \sum_{t} & || x^{eef}_{t} - x^{eef}_{t+1} ||_{2} + || \psi^{eef}_{t} - \psi^{eef}_{t} ||^{2} \\
        & + || x^{eef}_{t_{g}} - p^{eef}_{t_{g}} || 
    \end{split}
    \end{equation*}
    Additionally, a trajectory is chosen if the position of the wrist is within the threshold $\beta < 0.15m$ at the point of contact. 
    \item \texttt{Deep Squat \& Pick:} Stability is enforced by introducing cost terms that minimize foot slippage, including penalties on foot velocity and on changes in inter-foot distance. 
    To further improve the grip on the object we add the following cost. 
    \[ q^{ref} = argmin ||  d_{eef} - \lambda_{eef} ||_{2} \].  
    This cost minimizes the distance $(d_{eef})$ between the right and left hand end-effectors to match the width of the box $\lambda_{eef}$.
    A filtering step prior to optimization is performed, we eliminate all trajectories that have yaw rotation $|\psi_{root}| > \pi/6 $ this would lead to a distribution of trajectories facing forward. 
    \item \texttt{Pour:} The obstacle avoidance and filtering process of pour is similar to that of pick. Additionally we perform the following optimization on the wrist to pour rotate the wrist. 
\begin{equation*}
\begin{split}
    q^{ref} = \arg\min \sum_{t} & || x^{eef}_{t} - x^{eef}_{t+1} ||_{2} + d(x^{eef}, \lambda) + \\
    & || \psi^{eef} - \psi^{eef}_{des} ||^{2}
\end{split}
\end{equation*}
    \item \texttt{Vertical Wipe:} A cost is added to to track the control trajectory $\tau_{wrist}$. 
    \item \texttt{Deep Squat:} Similar to Bimanual Pick, we add costs to minimize feet slippage and for better stability. 
\end{itemize}

\subsection{RL Policy Details}

The task specific rewards used in individual tasks is given in Table.~\ref{tab:task_rewards}, 
additionally the policy uses motion tracking rewards. 
The \texttt{Deep Squat} and \texttt{Bimanual Pick} have rewards to minimize root planar linear and angular velocity. 
The actor and critic networks are initialized by MLPs.

\end{document}